\documentclass[sigconf]{acmart}

\usepackage{multirow}
\usepackage{algorithm}
\usepackage[noend]{algorithmic}
\usepackage{subcaption}
\usepackage{float}
\usepackage{enumitem}
\newtheorem{theorem}{Theorem}

\newtheorem{definition}{Definition}

\AtBeginDocument{%
  }

\copyrightyear{2024}
\acmYear{2024}
\setcopyright{acmlicensed}
\acmConference[KDD '24] {Proceedings of the 30th ACM SIGKDD Conference on Knowledge Discovery and Data Mining }{August 25--29, 2024}{Barcelona, Spain.}
\acmBooktitle{Proceedings of the 30th ACM SIGKDD Conference on Knowledge Discovery and Data Mining (KDD '24), August 25--29, 2024, Barcelona, Spain}
\acmISBN{979-8-4007-0490-1/24/08}
\acmDOI{10.1145/3637528.3672353}

\begin{document}

\title{Decision Focused Causal Learning for Direct Counterfactual Marketing Optimization}

\author{Hao Zhou}
\authornote{Both authors contributed equally to this research.}
\orcid{0009-0008-5204-663X}
\affiliation{%
  \institution{State Key Laboratory for Novel Software Technology}
  \institution{Nanjing University}
  \city{Nanjing}
  \country{China}
}
\affiliation{%
  \institution{Meituan}
  \city{Beijing}
  \country{China}
}
\email{zhouhao29@meituan.com}

\author{Rongxiao Huang}
\authornotemark[1]
\orcid{0009-0000-7140-5601}
\affiliation{%
  \institution{State Key Laboratory for Novel Software Technology}
  \institution{Nanjing University}
  \city{Nanjing}
  \country{China}
}
\email{rxhuang@smail.nju.edu.cn}

\author{Shaoming Li}
\orcid{0000-0002-4915-9958}
\affiliation{%
  \institution{Meituan}
  \city{Beijing}
  \country{China}
}
\email{shaoming.li@outlook.com }

\author{Guibin Jiang}
\orcid{0009-0008-2558-4299}
\affiliation{%
  \institution{Meituan}
  \city{Beijing}
  \country{China}
}
\email{jiangguibin@meituan.com}

\author{Jiaqi Zheng}
\orcid{0000-0001-8403-9655}
\authornote{Corresponding author.}
\affiliation{%
  \institution{State Key Laboratory for Novel Software Technology}
  \institution{Nanjing University}
  \city{Nanjing}
  \country{China}
}
\email{jzheng@nju.edu.cn}

\author{Bing Cheng}
\orcid{0009-0007-8390-8717}
\affiliation{%
  \institution{Meituan}
  \city{Beijing}
  \country{China}
}
\email{bing.cheng@meituan.com}

\author{Wei Lin}
\orcid{0000-0003-2851-820X}
\affiliation{
  \institution{Meituan}
  \city{Beijing}
  \country{China}
}
\email{lwsaviola@163.com}

\renewcommand{\shortauthors}{Hao Zhou et al.}

\begin{abstract}
Marketing optimization plays an important role to enhance user engagement in online Internet platforms. Existing studies usually formulate this problem as a budget allocation problem and solve it by utilizing two fully decoupled stages, i.e., machine learning (ML) and operation research (OR). However, the learning objective in ML does not take account of the downstream optimization task in OR, which causes that the prediction accuracy in ML may be not positively related to the decision quality.

Decision Focused Learning (DFL) integrates ML and OR into an end-to-end framework, which takes the objective of the downstream task as the decision loss function and guarantees the consistency of the optimization direction between ML and OR. However, deploying DFL in marketing is non-trivial due to multiple technological challenges. Firstly, the budget allocation problem in marketing is a 0-1 integer stochastic programming problem and the budget is uncertain and fluctuates a lot in real-world settings, which is beyond the general problem background in DFL. Secondly, the counterfactual in marketing causes that the decision loss cannot be directly computed and the optimal solution can never be obtained, both of which disable the common gradient-estimation approaches in DFL. Thirdly, the OR solver is called frequently to compute the decision loss during model training in DFL, which produces huge computational cost and cannot support large-scale training data. In this paper, we propose a decision focused causal learning framework (DFCL) for direct counterfactual marketing optimization, which overcomes the above technological challenges. Both offline experiments and online A/B testing demonstrate the effectiveness of DFCL over the state-of-the-art methods. Currently, DFCL has been deployed in several marketing scenarios in Meituan, one of the largest online food delivery platform in the world. 

\end{abstract}

\ccsdesc[500]{Computing methodologies~Machine learning approaches}
\ccsdesc[500]{Applied computing~Electronic commerce}
\keywords{Causal Inference, Decision Focused Learning, Marketing Optimization}



\maketitle

\section{Introduction}
Conducting marketing campaigns is a popular and effective way used by online Internet platforms to boost user engagement and revenue. For example, coupons in Taobao\cite{zhang2021bcorle} can stimulate user activity, dynamic pricing in Airbnb\cite{ye2018customized} and discounts in Uber\cite{du2019improve} encourage users to use the products. 

Despite the incremental revenues, marketing campaigns could incur significant costs. In order to be sustainable, a marketing campaign is usually conducted under a limited budget. In other words, only a portion of individuals (e.g., shops or goods) may receive marketing treatments due to a limited budget. Hence, assigning the appropriate marketing treatments to different individuals is essential for the effectiveness of a marketing campaign since users would respond differently to various promotional offers. Such decision problems can be formalized as resource allocation problems and have been investigated for decades.

The mainstream solution for these problems is a two-stage method \cite{ai2022lbcf,zhao2019unified,albert2022commerce,wang2023multi, du2019improve}. In the first stage, the individual-level (incremental) response under different treatments is predicted using ML models. The second stage is OR, and the predictions are fed into the combination optimization algorithms to achieve optimal overall revenue. However, the objectives of the two stages are not aligned: the former focuses on the predictive precision of the ML models, while the latter focuses on the quality of decisions. The method has some defects due to the isolation of ML and OR. First, the prediction precision of ML models has no strict positive correlation with the quality of the final decision. This is because standard loss functions (e.g., mean square error, cross-entropy error) do not take the interplay between the predictions into account, which can affect decision quality. Second, ML models often fall short of perfect precision, and the complex operations performed on the predictions in OR lead to the amplification or accumulation of prediction errors. Thus, the two-stage method usually obtains suboptimal decisions and is even inferior to heuristic strategies in some scenarios. 

Recently, Decision-Focused Learning (DFL) \cite{mandi2023decision, amos2017optnet, poganvcic2019differentiation, elmachtoub2022smart} has received increasing attention as an appropriate alternative to the two-stage method. The paradigm integrates prediction and optimization into an end-to-end system, which effectively aligns the objectives of both stages and achieves better performance on many challenging tasks. The key idea is to train ML models using a loss function that directly measures the quality of the decisions obtained from the predictions. Specifically, the ML models are trained under the predict-then-optimize framework \cite{elmachtoub2022smart}, which (1) makes predictions based on historical data, (2) solves the optimization problem based on the predictions, and (3) computes the decision loss to update the ML model parameters using stochastic gradient descent (SGD). 

Nevertheless, deploying DFL in marketing is non-trivial due to the following challenges.

\textbf{Uncertainty of constraints.} Most prior works of DFL have investigated the optimization problem where the unknown parameters appear in the objective function. The reason behind this is that the unknown parameters in the constraints lead to uncertainty in the solution space, and the optimal solution derived from the predictions may not be feasible under the real parameters. Within the constraints of our optimization problem, there are two distinct forms of uncertainty: intrinsic and extrinsic. The inherent uncertainty in the constraints refers to the costs consumed by the individuals under different treatments, which can be predicted based on historical data. Extrinsic uncertainty is the frequently changing marketing budget, determined by the external environment. An ML model is required to guarantee superior performance under different marketing budgets. Thus, our optimization objective is the effectiveness of the decision under any budget, and the optimization problem is a 0-1 integer stochastic programming.

\textbf{Counterfactuals in marketing.} Computing decision loss in marketing is challenging due to the presence of counterfactuals. Specifically, observing the values and costs of an individual under different treatments is impossible because the individual can only receive one treatment, which is also called the fundamental problem of causal inference~\cite{sekhon2008neyman}. In addition, the optimal solution of the optimization problem cannot be obtained based on offline data due to the counterfactuals, which disables the common gradient-estimation methods (e.g., SPO~\cite{elmachtoub2022smart}, LODL~\cite{shah2022decision}, LTR~\cite{mandi2022decision}) in DFL.

\textbf{Computational cost of large-scale dataset.} Computational cost is one of the major roadblocks for DFL involving large-scale optimization. As mentioned above, DFL integrates prediction and optimization into an end-to-end system, where the solver will be called frequently during training to solve the optimization problem. Therefore, the computational cost of DFL is high, leading prior works to investigate toy-level problems with few decision variables. In real-world applications, we need to train models for tens of millions of data, which is unsupportable by traditional DFL.

In this paper, we propose Decision-Focused Causal-Learning (DFCL) to address the above challenges. The main contributions of this work can be summarized as follows.

\textbf{Generalization.} In order to address both endogenous uncertainty (cost of individual consumption) and exogenous uncertainty (marketing budget) in the constraints, the uncertainty constraints are transformed into the objective function of the dual problem using Lagrangian duality theory. The optimization objective of the dual problem is then used as the decision loss. Moreover, we prove that the budget of the primal problem corresponds to the Lagrange multipliers of the dual problem, and thus optimizing the dual solution under different Lagrange multipliers is equivalent to optimizing the quality of decisions under different budgets.

\textbf{Counterfactual Decision Loss.} Optimal solution, decision loss, and gradient cannot be computed directly due to the existence of counterfactuals in marketing, thus we propose two solutions: (1) surrogate loss function and (2) black-box optimization based on the Expected Outcome Metric (EOM) \cite{ai2022lbcf, zhou2023direct, zhao2017uplift}. Inspired by Policy Gradient in Reinforcement Learning, we transform the decision problem of discrete actions into the problem which maximizes expected revenue under the probability distribution of the actions, and combine the Maximum Entropy Regularizer as well as the Lagrangian duality theory to give two kinds of surrogate loss functions: Policy Learning Loss and Maximun Entropy Regularized Loss. We theoretically guarantee continuity, convexity and equivalence of the surrogate loss functions. For black-box optimization, we employ the EOM to give an unbiased estimation of the decision loss and improve the finite difference strategy to develop an efficient estimator of the gradient, which enables us to update the model parameters using gradient descent.

\textbf{Scalability.} In real-world applications, we need to train models for tens-of-millions of data. The surrogate functions proposed in this paper are smooth convex loss functions with almost the same computational efficiency as the two-stage method. For black-box optimization, frequently solving the optimization problem after perturbation incurs huge computational overhead. We accelerate the problem solving and modify the gradient estimator using the Lagrangian duality theory, which significantly improves the training efficiency and reduces the training time from hour-level to second-level per epoch compared to the black-box method based on the primal problem.

We conduct extensive experiments to evaluate the performance of DFCL. Both offline experiments and online A/B testing show the superior performance of our method over state-of-the-art baselines. DFCL is deployed to several scenarios in Meituan, an online food delivery platform, and achieves significant revenue.
\section{Related Works}

\textbf{Two-stage Method.} The mainstream solution to the resource allocation problem in marketing usually follows the two-stage paradigm \cite{albert2022commerce, wang2023multi, zhao2019unified, ai2022lbcf}, which handles the two stages—machine learning (ML) and operation research (OR)—independently. In the first stage, the uplift models are deployed to predict the treatment effects of individuals. Some prior works have focused on the design of uplift models, including Meta-Learners \cite{kunzel2019metalearners, nie2021quasi}, Causal Forests \cite{athey2019generalized, wager2018estimation, zhao2017uplift, ai2022lbcf}, representation learning \cite{johansson2016learning, yao2018representation, shi2019adapting} and rank model \cite{betlei2021uplift, kuusisto2014support}. However, standard loss functions (such as mean square error and cross-entropy error) for training uplift models do not take the downstream OR into account. In the second stage, the resource allocation problem is represented as a multi-choice knapsack problem (MCKP), which is NP-Hard and efficiently solved based on Lagrangian duality theory \cite{ai2022lbcf, albert2022commerce, wang2023multi, zhou2023direct}.

\noindent \textbf{Decision-Focused Learning(DFL).} DFL is considered an appropriate alternative to the two-stage method, which integrates prediction and optimization into an end-to-end system. Since computing decision loss requires solving optimization problems, which usually involve non-differentiable operations, automatic differentiation in machine learning frameworks (such as Pytorch \cite{paszke2019pytorch} and Tensorflow \cite{abadi2016tensorflow}) cannot give the correct gradient. Three categories of approaches to gradient computation are proposed by prior DFL works: analytical smoothing of optimization mappings, smoothing by random perturbations, and differentiation of surrogate loss function. The first method derives the analytic gradient of decision loss by using the KKT condition or the homogenous self-dual formulation, including Optnet \cite{amos2017optnet}, DQP \cite{donti2017task}, QPTL \cite{wilder2019melding}, and IntOpt \cite{mandi2020interior}. However, when the optimization problem is discrete, the method requires a continuous relaxation of the primal problem, which results in suboptimality. A potential resolution is to consider every optimization problem as a black-box optimization and utilize random perturbations, such as DBB \cite{poganvcic2019differentiation}, DPO \cite{berthet2020learning}, and I-MLE \cite{niepert2021implicit}, to generate approximate gradient. Furthermore, the decision loss is typically discontinuous and nonconvex, so some of these works suggest convex surrogate functions, including SPO \cite{elmachtoub2022smart}, LTR \cite{mandi2022decision}, NCE \cite{mulamba2020contrastive}, and LODL \cite{shah2022decision}, for the decision loss. 

\noindent The most related works to ours are DRP \cite{du2019improve} and DPM \cite{zhou2023direct}. DRP proposes to directly learn ROI (ratio between incremental values and incremental costs) to rank and choose individuals in the binary treatment setting. It has been shown by \cite{zhou2023direct} that the loss function in DRP is unable to converging to a stable extreme point. DPM extend the idea to the multiple treatments setting by directly learning the unbiased estimation of the decision factor in OR. However, the construction of the decision factor in multi-treatment setting relies on the law of diminishing marginal utility, which does not hold strictly in some scenarios of marketing.
\section{Problem Formulation}
\label{sec:formulation}

In this section, we formalize the resource allocation problem and introduce the overall optimization objective in marketing. 



We start with a common marketing scenario that has $M$ types of treatments. Let $r_{ij}$ and $c_{ij}$ be the revenue and cost of individual $i$ under treatment $j$, respectively. The objective is to find an optimal allocation strategy for a group of individuals to maximize the revenue of the platform, given a limited budget $B$. Therefore, the budget allocation problem with multiple treatments (MTBAP) can be formulated as an integer programming problem~\eqref{Eq:MTBAP}:

\begin{equation}
\label{Eq:MTBAP}
\begin{aligned}
\max_z \ \ F(z,B) \ \ = \ \ &\sum_{i}\sum_{j} z_{ij} r_{ij},\\
\text {s.t.} \ \ &\sum_{i}\sum_{j} z_{ij} c_{ij} \le B,\\
&\sum_{j} z_{ij} = 1, \forall i,\\
&z_{ij} \in \{0,1\},\forall i,j, 
\end{aligned}
\end{equation}
where $z_{ij} \in \{0,1\}$ is the decision variable to denote whether to assign treatment $j$ to individual $i$. The first constraint is the limitation of the budget and the second one requires that only one treatment is assigned to each individual. Since the budget $B$ fluctuates a lot in real-world settings, the objective is regared as a function of the budget and the overall marketing goal is to maximize revenue $F(z,B)$ within arbitrary given budget.

\noindent \textbf{Combinatorial Optimization Algorithm.}  When the value of $r_{ij}$ and $c_{ij}$ are known in advance, MTBAP is a classical multiple choice knapsack problem (MCKP)~\cite{sinha1979multiple}, which remains NP-Hard.  Existing studies usually solve this problem by using greedy algorithms or Lagrangian duality theory, both of which can provide a approximation ratio of 
$$\rho = 1 - \frac{\max_{ij} r_{ij}}{\mathrm{OPT}},$$
where $\mathrm{OPT}$ is the optimal solution. In the above equation, $\max_{ij} r_{ij}$ refers to the revenue of one individual (e.g., one user or one shop), which is negligible compared with $\mathrm{OPT}$ that is the sum of the revenue of all the individuals in marketing. Therefore, it indicates that both greedy algorithms and Lagrangian duality theory can achieve near optimal performance, which are also the most common algorithms to solve MTBAP in marketing. The details can be found in existing works, which will not be discussed in this paper.

\noindent \textbf{Model Prediction.} However, the value of $r_{ij}$ and $c_{ij}$ are unknown during decision making in real-world applications, which are usually replaced with the prediction value. Therefore, how to make the prediction of $r_{ij}$ and $c_{ij}$ plays important roles in marketing effectiveness, which will be addressed in this paper. In the traditional two-stage approaches, the machine learning (ML) model is trained with the direction of optimizing prediction accuracy, which may be not consistent with the direction of optimizing decision quality.
In the following sections, we mainly focus on the design of the loss function, to make a tradeoff between the prediction accuracy and the decision quality.

\section{Learning Framework of DFCL}
\label{sec:loss}

In the learning framework, the loss function includes two parts, the prediction loss and the decision loss, i.e.,
$$\mathcal{L}_{DFCL}=\alpha \mathcal{L}_{PL}+ \mathcal{L}_{DL}.$$
The former $\mathcal{L}_{PL}$ aims to decrease the prediction error, which contributes to improving the generalization ability of a ML model. The latter $\mathcal{L}_{DL}$ measures the decision quality of the downstream task, which is exactly the objective of marketing optimization.

\subsection{Prediction Loss}
\label{subsec:prediction}

In the traditional two-stage method, the ML model is trained by minimizing the difference between the predictions $\hat{r},\hat{c}$ and the ground-truth values $r,c$. For instance, in a regression problem, the mean squared error (MSE) is usually used to train the ML model:
\begin{equation}\label{eq:L_MSE}
     \mathcal{L}_{MSE}(r,c,\hat{r},\hat{c}) = \frac{1}{NM} \sum_{i} \sum_{j} (r_{ij}-\hat{r}_{ij})^{2}+(c_{ij}-\hat{c}_{ij})^{2}.
\end{equation}
Due to the counterfactuals in marketing, observing the revenue or cost of an individual under different treatments is impossible because each individual can only receive one treatment, which is also called the fundamental problem of causal inference. 

\begin{definition}[The fundamental problem of causal inference]
For all individuals, only one of all the potential outcomes under different treatments can be observed in real-world data.
\end{definition}

Therefore, $\mathcal{L}_{MSE}$ cannot be directly computed according to Eq.~\ref{eq:L_MSE} since $r_{ij_1}$ and $r_{ij_2}$ (or equivalently, $c_{ij_1}$ and $c_{ij_2}$) cannot be simultaneously observed for any $j_1 \neq j_2$. To solve this problem, we first formulate the training data set and then develop a equivalent prediction loss in marketing.

\noindent\textbf{Data Set.} Suppose that there is a data set of size $N$ collected from random control trials (RCT). The $i$-th sample is denoted by $(x_i, t_i, r_{it_i}, c_{it_i})$, where $x_i$ is the features of individual $i$, $t_i$ is the assigned treatment, and $r_{it_i}, c_{it_i}$ are the revenue and the cost of individual $i$ under treatment $t_i$. Denote the count of the samples (individuals) receiving treatment $j$ by $N_j$. 

\noindent\textbf{Prediction Loss.} Given the above data set, we present the prediction loss in marketing in Eq.~\eqref{Eq:PL}. Theorem~\ref{theorem:L_PL} presents the equivalency and the detailed proof can be found in Appendix~\ref{sec:proof_L_PL}.

\begin{equation}\label{Eq:PL}
     \mathcal{L}_{PL}(r,c,\hat{r},\hat{c})  = \frac{1}{M}\sum_{i}\frac{1}{N_{t_i}} [(r_{it_{i}}-\hat{r}_{it_{i}})^{2}+(c_{it_{i}}-\hat{c}_{it_{i}})^{2}].
\end{equation}

\begin{theorem}\label{theorem:L_PL}
The prediction loss $\mathcal{L}_{PL}$ is equivalent to $\mathcal{L}_{MSE}$, i.e.,
$$\mathcal{L}_{PL}=\mathcal{L}_{MSE}.$$
\end{theorem}

\subsection{Decision Loss}
\label{subsec:decision}
As is stated in Sec.~\ref{sec:formulation}, the ground-truth value of $r$ and $c$ are usually unknown in advance, which are replaced with the prediction $\hat{r}$ and $\hat{c}$ during decision making. Therefore, denote the original optimization problem $F(z,B)$ by $F(z,B,\hat{r},\hat{c})$, and the solution $z^*(B,\hat{r},\hat{c})$ is obtained by solve MTBAP  $F(z,B,\hat{r},\hat{c})$, i.e.,
$$z^*(B,\hat{r},\hat{c}) = \arg\max_z F(z,B,\hat{r},\hat{c}).$$
The objective value achieved by the current solution $z^*(B,\hat{r},\hat{c})$ can be expressed with the ground-truth value of $r$ as
$$\sum_i\sum_j r_{ij} z^*_{ij}(B,\hat{r},\hat{c}).$$

The decision loss under budget $B$ is defined as the negative of the objective value with ground-truth $r$ and predicted decision $z^{*}(B, \hat{r}, \hat{c})$, i.e.,
$$\mathcal{L}_{DL} (B,r,c,\hat{r},\hat{c}) = - \sum_i\sum_j r_{ij} z^*_{ij}(B,\hat{r},\hat{c}).$$
As is descripted in Sec.~\ref{sec:formulation}, the budget $B$ fluctuates a lot in real-world settings and the overall marketing objective is to maximize the revenue under arbitrary budget. Therefore, the decision loss in marketing is defined as
\begin{align*}
\mathcal{L}_{DL}(r,c,\hat{r},\hat{c}) &= \int_{0}^{\infty} \mathcal{L}_{DL} (B,r,c,\hat{r},\hat{c}) dB \\
&= \int_{0}^{\infty} - \sum_i\sum_j r_{ij} z^*_{ij}(B,\hat{r},\hat{c}) dB.
\end{align*}
For ease of calculation, we can also discretize the budget and compute the decision loss by
$$\mathcal{L}_{DL}(r,c,\hat{r},\hat{c}) = \sum_{B} \mathcal{L}_{DL} (B,r,c,\hat{r},\hat{c}).$$

\subsection{Learning Framework}

Algorithm~\ref{alg:DL} presents the framework of decision focused causal learning (DFCL). The most crucial step in this framework is the gradient estimation of $\mathcal{L}_{DFCL}$ in line~10 of Algorithm~\ref{alg:DL}. However, it is non-trival in marketing due to the following technological challenges, i.e., uncertainty of constraints, counterfactual and computation cost.
In this paper, we will show how to address these challenges and how to deploy DFCL in marketing optimization.

\begin{algorithm}
	\renewcommand{\algorithmicrequire}{\textbf{Input:}}
	\renewcommand{\algorithmicensure}{\textbf{Output:}}
	\caption{Decision Focused Causal Learning (DFCL)}
	\label{alg:DL}
	\begin{algorithmic}[1]
		\REQUIRE training data D $\equiv \{(x_{i}, t_{i}, r_{it_{i}}, c_{it_{i}})\}^{N}_{i=1}$
  
		\STATE Initialize $\omega$
		\FOR{each epoch}
            \STATE $\hat{r}, \hat{c} = m_{\omega}(x).$
            \STATE $ \mathcal{L}_{PL} = \sum_{i}\frac{1}{N_{t_i}} [(r_{it_{i}}-\hat{r}_{it_{i}})^{2}+(c_{it_{i}}-\hat{c}_{it_{i}})^{2}].$
            \FOR{each budget $B$}
                \STATE $z^{*}(B, \hat{r}, \hat{c})  = \arg\max_z F(z,B,\hat{r}, \hat{c})$.
                \STATE $\mathcal{L}_{DL} (B) = - \sum_i\sum_j r_{ij} z^*_{ij}(B,\hat{r},\hat{c}).$
            \ENDFOR
		\STATE $\mathcal{L}_{DL} = \sum_{B} \mathcal{L}_{DL} (B)$
            \STATE $\mathcal{L}_{DFCL}=\alpha \mathcal{L}_{PL}+ \mathcal{L}_{DL}.$
		\STATE $\omega = \omega - \eta \frac{\partial{\mathcal{L}_{DFCL}}}{\partial{\omega}}$
		\ENDFOR
	\end{algorithmic}  
\end{algorithm}
\section{Gradient Estimation of DFCL}
\label{sec:gradient}

The loss of DFCL consists of the prediction loss and the decision loss. The former is a continuously differentiable function whose gradient can be directly computed. Hence, the gradient estimation of the decision loss is the key focus of this section. Firstly, we introduce the equivalent dual decision loss to remove the uncertain constraints and reduce the computation cost of combinatorial optimization algorithms. Secondly, we develop two surrogate loss functions and improve the black-box optimization algorithm  to provide a gradient estimation of the dual decision loss. 

\subsection{Dual Decision Loss}
\label{subsec:dual}

Based on the Lagrangian duality theory, the upper bound of the original problem $F(z,B,r,c)$ can be obtained by solving the following dual problem~\eqref{Eq:MTBAP-Dual}.
\begin{align}
&\min_{\lambda \ge 0} \left (
\begin{array}{c}
\max\limits_z \lambda B + \sum_{i}\sum_j (r_{ij} - \lambda c_{ij})z_{ij}  \\
s.t. \ \sum_j z_{ij} = 1, \forall j \\
z_{ij} \in \{0,1\}, \forall i, j
\end{array}
\right ) \nonumber\\
= & \min_{\lambda \ge 0} \max_z H(z,\lambda, B,r,c) \nonumber \\
= & \min_{\lambda \ge 0} G(\lambda, B,r,c). \label{Eq:MTBAP-Dual}
\end{align}
The optimal Lagrange multiplier $\lambda^*$ for the dual problem~\eqref{Eq:MTBAP-Dual} can be obtained by using a gradient descent algorithm or a binary search method with the terminal condition of $B - \sum_i\sum_j c_{ij}z_{ij} \le \epsilon$ or $\lambda \le \epsilon$. In addition, an approximately optimal solution for the original problem can be derived by maximizing $H(z,\lambda^*,B,r,c)$. Theorem~\ref{theorem:Dual} presents the relationship between the original problem $F(z,B,r,c)$ and the dual problem $G(\lambda, B,r,c)$.

\begin{theorem}\label{theorem:Dual}
Denote by $F_c(z,B,r,c)$ the relaxation form of $F(z,B,r,c)$ where the decision variables $z$ are relaxed to continuous variables (i.e., $z_{ij}\in [0,1]$ for $\forall i,j$). Denote the optimal solution by 
\begin{align*}
z_c^*(B,r,c) &= \arg\max_z F_c(z,B,r,c), \nonumber\\
z^*(B,r,c) &= \arg\max_z F(z,B,r,c), \nonumber\\
\lambda^*(B,r,c) &= \arg\min_{\lambda\ge 0} G(\lambda,B,r,c). \nonumber
\end{align*}
Given the optimal  Lagrange multiplier $\lambda^*$, an approximation solution for the original problem can be derived by
\begin{equation*}
z^d(\lambda^*, B,r,c) = \arg\max_z H(z,\lambda^*,B,r,c).
\end{equation*}
Based on these definitions, we claim that $\lambda^*$ is monotonic decreasing with the increment of the budget $B$, and we have 
\begin{align*}
F(z^d, B,r,c) &\le F(z^*, B,r,c) \\
&\le F_c(z_c^*,B,r,c) \\
&= G(\lambda^*, B,r,c) \\
&\le F(z^d, B,r,c) + \max_{ij} r_{ij}
\end{align*}
\end{theorem}

The detailed proof can be found in~\cite{kellerer2004multiple}. Given the optimal $\lambda^*$, Theorem~\ref{theorem:Dual} indicates that the solution $z_d(\lambda^*, B,r,c)$ obtained by maximizing $ H(z,\lambda^*,B,r,c)$ is approximately optimal with an appriximation ratio of
\begin{align*}
\rho = \frac{F(z^d, B,r,c)}{F(z^*, B,r,c)} &\ge \frac{F(z^*, B,r,c) - \max_{ij} r_{ij}}{F(z^*, B,r,c)}  \\
& = 1 - \frac{\max_{ij} r_{ij}}{{F(z^*, B,r,c)}} \\
&\approx 1
\end{align*}
The last equality holds because $F(z^*, B,r,c)$ is the sum of the revenue of millions of individuals in marketing, which means that $F(z^*, B,r,c) \gg \max_{ij} r_{ij}$. 

Therefore, instead of the original problem, the optimization of the dual problem $H(z,\lambda^*,B,r,c)$ is taken as the learning objective, which we call the dual decision loss. Given the optimal $\lambda^*$ and the prediction value $\hat{r},\hat{c}$, the solution $z^d(\lambda^*, B,\hat{r},\hat{c})$ is obtained by maximizing $H(z,\lambda^*,B,\hat{r},\hat{c})$, i.e.,
\begin{equation*}
z^d(\lambda^*, B,\hat{r},\hat{c}) = \arg \max_z H(z,\lambda^*,B,\hat{r},\hat{c}).
\end{equation*}
Notice that $\lambda^*B$ can be taken as a constant, and removing it from $H(z,\lambda^*,B,\hat{r},\hat{c})$ does not influence $z^d$. Therefore, the solution $z^d$ can be rewritten as 
\begin{equation*}
z^d(\lambda^*, \hat{r},\hat{c}) = \arg \max_z H(z,\lambda^*,\hat{r},\hat{c}),
\end{equation*}
where $H(z,\lambda^*,\hat{r},\hat{c})$ is the form of $H(z,\lambda^*,B,\hat{r},\hat{c})$ after removing $\lambda^*B$.
The dual decision loss achieved by the current solution $z_d$ is 
\begin{equation*}
\mathcal{L}_{DDL}(\lambda^*, B, r,c,\hat{r},\hat{c}) = - (\lambda^*B + \sum_i\sum_j (r_{ij} - \lambda^* c_{ij}) z^d_{ij}(\lambda^*, \hat{r},\hat{c})).
\end{equation*}
Similarly, since $\lambda^*$ and $B$ is irrelevant to the prediction value $\hat{r},\hat{c}$, $\lambda^*B$ can be regarded as a constant and removed from the dual decision loss. 
According to Theorem~\ref{theorem:Dual}, $\lambda^*$ is monotonic decreasing with the increment of the budget $B$ and there is an unique $\lambda^*$ for the dual problem when given the budget $B$. Therefore, the decision loss $\mathcal{L}_{DL}(r,c,\hat{r},\hat{c})$ in the original problem under arbitrary budget $B$ can be transformed to the dual decision loss $\mathcal{L}_{DDL}(r,c,\hat{r},\hat{c})$ under arbitrary  Lagrange multiplier $\lambda^*$, i.e.,
\begin{align*}
\mathcal{L}_{DDL}(r,c,\hat{r},\hat{c}) &=  \int_{0}^\infty \mathcal{L}_{DDL}(\lambda^*, r,c,\hat{r},\hat{c}) d\lambda^* \\
&=  \int_{0}^\infty \mathcal{L}_{DDL}(\lambda, r,c,\hat{r},\hat{c}) d\lambda \\
&= - \int_{0}^\infty \sum_i\sum_j (r_{ij} - \lambda c_{ij}) z^d_{ij}(\lambda,\hat{r},\hat{c}) d\lambda.
\end{align*}
By discretizing the Lagrange multiplier $\lambda$, the dual decision loss can be computed by
$$\mathcal{L}_{DDL}(r,c,\hat{r},\hat{c}) = \sum_{\lambda} \mathcal{L}_{DDL} (\lambda,r,c,\hat{r},\hat{c}).$$

\subsection{Policy Learning Loss}
\label{subsec:policylearning}


Notice that the dual problem $H(z,\lambda,\hat{r},\hat{c})$ can be solved by 
\begin{align*}
 \max_z H(z,\lambda,\hat{r},\hat{c}) 
=& \left (
\begin{array}{c} 
\max\limits_z \sum_{i}\sum_j (\hat{r}_{ij} - \lambda \hat{c}_{ij})z_{ij}  \\
s.t. \ \sum_j z_{ij} = 1, \forall j \\
z_{ij} \in \{0,1\}, \forall i, j
\end{array}
\right ) \\
=& \sum_{i} \max_j (\hat{r}_{ij} - \lambda \hat{c}_{ij})
\end{align*}
Therefore, the solution $z^d(\lambda, \hat{r},\hat{c}) = \arg \max_z H(z,\lambda,\hat{r},\hat{c}) $ can be expressed by
$$z^d_{ij}(\lambda,\hat{r},\hat{c}) = \mathbb{I}_{j = \arg\max_j \hat{r}_{ij} - \lambda \hat{c}_{ij}}.$$
Hence, the dual decision loss is rewritten as
$$\mathcal{L}_{DDL}(r,c,\hat{r},\hat{c}) = -\sum_{\lambda} \sum_i\sum_j (r_{ij} - \lambda c_{ij}) \mathbb{I}_{j = \arg\max_j \hat{r}_{ij} - \lambda \hat{c}_{ij}}$$
However, $\mathcal{L}_{DDL}(r,c,\hat{r},\hat{c})$ is not differentiable with respect to $\hat{r}$ and $\hat{c}$ due to the indicator function. Instead, we utilize a softmax function to smooth the dual decision loss, i.e.,
\begin{equation}\label{Eq:L'_DDL}
\mathcal{L}'_{DDL}(r,c,\hat{r},\hat{c}) = -\sum_{\lambda} \sum_i\sum_j (r_{ij} - \lambda c_{ij}) \frac{\exp(\hat{r}_{ij} - \lambda \hat{c}_{ij})}{ \sum_k \exp(\hat{r}_{ik} - \lambda \hat{c}_{ik})}
\end{equation}
Let $p_{ij}(\lambda,\hat{r},\hat{c}) =\exp(\hat{r}_{ij} - \lambda \hat{c}_{ij}) /  \sum_k \exp(\hat{r}_{ik} - \lambda \hat{c}_{ik})$ be the probability of assigning treatment $j$ to individual $i$. Take $r_{ij} - \lambda c_{ij}$ as the reward of assigning treatment $j$ to individual $i$.
Hence, minimizing $\mathcal{L}'_{DDL}(r,c,\hat{r},\hat{c})$ is equivalent to maximizing the expected reward of policy $\pi = p_{ij}(\lambda,\hat{r},\hat{c})$ under different Lagrange multipliers. Therefore, $\mathcal{L}'_{DDL}$ is also called the policy learning loss.

Due to the counterfactual in marketing, $\mathcal{L}'_{DDL}(r,c,\hat{r},\hat{c})$ cannot be directly computed by Eq.~\eqref{Eq:L'_DDL} in training data sets. Instead, we propose a surrogate  loss, i.e.,
$$\mathcal{L}_{PLL}(r,c,\hat{r},\hat{c})=-\sum_{\lambda} \sum_{i}\frac{N}{N_{t_i}}(r_{it_{i}}-\lambda c_{it_{i}})\frac{\exp(\hat{r}_{it_i}-\lambda\hat{c}_{it_i})}{\sum_{j}\exp(\hat{r}_{ij}-\lambda\hat{c}_{ij})}.$$
Theorem~\ref{theorem:L_PLL} presents the equivalence between the original dual decision loss $\mathcal{L}_{DDL}$ and the surrogate policy learning loss $L_{PLL}$. The detailed proof can be found in Appendix~\ref{sec:proof_L_PLL}.

\begin{theorem}\label{theorem:L_PLL}
$\mathcal{L}_{DDL},\mathcal{L}'_{DDL}$ and $\mathcal{L}_{PPL}$ are equivalent, i.e.,
$$\mathcal{L}_{PLL}(\lambda,\hat{r},\hat{c}) = \mathcal{L}'_{DDL}(\lambda,\hat{r},\hat{c})$$
and
$$\min_{\hat{r},\hat{c}} \mathcal{L}_{PLL}(\lambda,\hat{r},\hat{c}) = \min_{\hat{r},\hat{c}} \mathcal{L}_{DDL}(\lambda,\hat{r},\hat{c}).$$
\end{theorem}

\subsection{Maximum Entropy Regularized Loss}
\label{subsec:entropy}


In order to obtain a differentiable closed form of $z^d(\lambda,\hat{r},\hat{c})$ with respect to $\hat{r}$ and $\hat{c}$, we relax the discrete constraint $z \in \{0, 1\}$ to a continuous one $x \in [0, 1]$ and add a maximum entropy regularizer to the objective function in $H(z,\lambda,\hat{r},\hat{c})$. Hence, $H(z,\lambda,\hat{r},\hat{c})$ is transformed to a nonlinear convex function, i.e.,
\begin{equation*}
\begin{aligned}
\max_z \ \sum_{i}\sum_{j} (\hat{r}_{ij} &-\lambda \hat{c}_{ij}) z_{ij} - \tau\sum_{i}\sum_{j} z_{ij}\ln z_{ij} ,\\
s.t. \ \ &\sum_{j} z_{ij} = 1, \forall i, \\
&z_{ij} \in [0,1],
\end{aligned}
\end{equation*}
where $\tau$ denotes the penalty weight. The Lagrange relaxation function can be further rewritten as
\begin{equation*}
     L(z, \beta)= \sum_{i=1}^{N}\sum_{j=1}^{M} (r_{ij}-\lambda c_{ij})z_{ij} - \tau\sum_{i=1}^{N}\sum_{j=1}^{M} z_{ij}\ln z_{ij} - \sum_{i} \beta_{i}(1-\sum_{j}z_{ij}),
\end{equation*}
where $\beta$ is the dual variables on the equality constraint. When $\frac{\partial L(z,\beta)}{\partial z} = 0$ and $\frac{\partial L(z,\beta)}{\partial \beta} = 0$, the optimal solution is obtained by
$$z^d_{ij}=\frac{\exp [(\hat{r}_{ij}-\lambda \hat{c}_{ij})/\tau]}{\sum_{k} \exp[(\hat{r}_{ik}-\lambda \hat{c}_{ik})/\tau]},$$
which is continuously differentiable with respect to $\hat{r}$ and $\hat{c}$. Hence, the dual decision loss can be rewritten as
\begin{equation*}\label{Eq:L''_DDL}
\mathcal{L}''_{DDL}(r,c,\hat{r},\hat{c}) = -\sum_{\lambda} \sum_i\sum_j (r_{ij} - \lambda c_{ij}) \frac{\exp [(\hat{r}_{ij}-\lambda \hat{c}_{ij})/\tau]}{\sum_{k} \exp[(\hat{r}_{ik}-\lambda \hat{c}_{ik})/\tau]}
\end{equation*}

Similarly, $\mathcal{L}''_{DDL}$ cannot be directly computed due to the counterfactual in marketing. We propose a surrogate loss $\mathcal{L}_{MERL}(r,c,\hat{r},\hat{c})$ as follows, which we call the maximum entropy regularized loss,
\begin{equation*}
\mathcal{L}_{MERL}(r,c,\hat{r},\hat{c})=-\sum_{\lambda} \sum_{i}\frac{N}{N_{t_i}}(r_{it_{i}}-\lambda c_{it_{i}})\frac{\exp[(\hat{r}_{it_i}-\lambda\hat{c}_{it_i})/\tau]}{\sum_{j}\exp[(\hat{r}_{ij}-\lambda\hat{c}_{ij})/\tau]}.
\end{equation*}
Notice that $\mathcal{L}_{PLL}$ is a special case of $\mathcal{L}_{MERL}$, where the solution $z^d$ can be regarded as a temperature softmax function in $\mathcal{L}_{MERL}$.

\subsection{Improved Finite-Difference Strategy}
\label{subsec:FD}

In addition to constructing surrogate loss functions, we can also use the Expected Outcome Metric (EOM)~\cite{ai2022lbcf, zhou2023direct, zhao2017uplift} to give an unbiased estimate of the decision loss and leverage black-box optimization for decision-focused learning. 

EOM is a commonly used method for offline strategy evaluation based on randomized dataset. Given a batch of $N$ random samples and model predictions $\hat{r}$ and $\hat{c}$, an arbitrary allocation strategy $z(\hat{r}, \hat{c})$ can be evaluated: (1) find the set of individuals whose received treatment is equal to the treatment in the allocation strategy $z(\hat{r}, \hat{c})$, (2) then empirically estimate their per capita revenue and per capita cost:
\begin{alignat*}{3}
\bar{r}(r, c, \hat{r}, \hat{c})=\frac{1}{N} \sum_{i} \frac{1}{p_{t_{i}}} r_{t_{i}} \mathbb{I}_{t_{i}=\arg\max_{j} \ z_{ij}},\\
\bar{c}(r, c, \hat{r}, \hat{c})=\frac{1}{N} \sum_{i} \frac{1}{p_{t_{i}}} c_{t_{i}} \mathbb{I}_{t_{i}=\arg\max_{j} \ z_{ij}},
\end{alignat*}
where $p_{t_{i}}$ denotes the probability that a treatment is equal to $t_{i}$ in the randomized dataset. For the primal MCKP with budget $B$, we can use binary search to empirically estimate the per capita revenue under a per capita budget $\frac{B}{N}$ as is shown in Appendix~\ref{sec:EOM}. 
Therefore, we can redefine the decision loss as follows:
\begin{equation*}
     \mathcal{L}_{DL}(r, c, \hat{r}, \hat{c})=-\sum_{B}\bar{r}(B, r, c, \hat{r}, \hat{c}).
\end{equation*}
Since the computation of $\bar{r}(B, r, c, \hat{r}, \hat{c})$ involves many nondifferentiable operations, we consider them as black-box functions and estimate the gradient by perturbation. Using the finite difference strategy, the gradient of the decision quality with respect to $\hat{r}_{ij}$ is estimated as:
\begin{equation*}
     \frac{\partial \mathcal{L}_{DL}(r, c, \hat{r}, \hat{c})}{\partial \hat{r}_{ij}} = \frac{\mathcal{L}_{DL}(r, c, \hat{r} + e_{ij}h, \hat{c})-\mathcal{L}_{DL}(r, c, \hat{r}, \hat{c})}{h},
\end{equation*}
where $h$ is a small constant, and $e^{ij} \in \{0,1\}^{N \times M}$ is a matrix where only the element in the i-th row and j-th column is 1, and all other elements are 0. The gradient term $\frac{\partial \mathcal{L}_{DL}(r, c, \hat{r}, \hat{c})}{\partial \hat{c}_{ij}}$ can be computed similarly. We estimate the gradient by perturbing the predictions one by one and obtain the gradient matrix $\frac{\partial \mathcal{L}_{DL}(r, c, \hat{r}, \hat{c})}{\partial \hat{r}}$ and $\frac{\partial \mathcal{L}_{DL}(r, c, \hat{r}, \hat{c})}{\partial \hat{c}} \in R^{N \times M}$. 
 Finally, we derive the following loss function, which is able to train the ML model via gradient descent:
\begin{equation*}
     \mathcal{L}_{FDL}(r, c, \hat{r}, \hat{c}) = \sum_{i}\sum_{j}\frac{\partial \mathcal{L}_{DL}(r, c, \hat{r}, \hat{c})}{\partial \hat{r}_{ij}}\hat{r}_{ij}+\frac{\partial \mathcal{L}_{DL}(r, c, \hat{r}, \hat{c})}{\partial \hat{c}_{ij}}\hat{c}_{ij}.
\end{equation*}

Since the perturbations are performed one by one, $\bar{r}(B, r, c, \hat{r}, \hat{c})$ requires frequent evaluation, leading to a considerable time complexity of black-box optimization based on the primal MCKP~\cite{yan2023end}. In practice, we find that the number of samples $N$ tends to be in the millions or even tens of millions, so the time consumption for a training epoch reaches the level of hours. A possible approach is to only perturb some of the samples by sampling, but this may incur the loss of much of the gradient information. 

Instead, we accelerate the problem solving and modify the gradient estimator by using Lagrangian duality theory. Since the budget $B$ in the primal MCKP corresponds one-to-one to the $\lambda$ in the duality problem, the dual decision loss can be redefined as:
\begin{equation*}
     \mathcal{L}_{DDL}(r, c, \hat{r}, \hat{c})=-\sum_{\lambda}\bar{r}(\lambda, r, c, \hat{r}, \hat{c}) - \lambda \bar{c}(\lambda, r, c, \hat{r}, \hat{c}).
\end{equation*}
Although we avoid solving the primal MCKP, it is still necessary to frequently evaluate the per capita revenue and per capita cost after perturbation under multiple Lagrangian multipliers. We observe that the decision making is independent for each individual thanks to the decomposition of the Lagrangian duality theory. Thus, for each sample, the smallest perturbation that causes a change in the dual decison loss is first calculated, and the loss after the perturbation is obtained by correcting only the original result. Appendix \ref{sec:estimator} provides details of the modified gradient estimator, which greatly reduces the computational overhead. Finally, the black-box optimization loss function can be rewritten as
\begin{equation*}
     \mathcal{L}_{IFDL}(r, c, \hat{r}, \hat{c}) = \sum_{i}\sum_{j}\frac{\partial \mathcal{L}_{DDL}(r, c, \hat{r}, \hat{c})}{\partial \hat{r}_{ij}}\hat{r}_{ij}+\frac{\partial \mathcal{L}_{DDL}(r, c, \hat{r}, \hat{c})}{\partial \hat{c}_{ij}}\hat{c}_{ij}.
\end{equation*}


It is sufficient to support model training on tens of million of data since the computational cost of incremental updating ${L}_{DDL}(r, c, \hat{r}, \hat{c})$ after perturbation is much less than that of re-evaluating it. To improve numerical stability in training, we truncate the perturbation matrix $P \in \mathbb{R}^{N \times M}$. Further, the loss function can be smoothed using Softmax to reduce the difficulty of training.
\begin{equation*}
     \mathcal{L}_{IFDL-Softmax}(r, c, \hat{r}, \hat{c}) = \sum_{i}\sum_{j}\frac{\partial \mathcal{L}_{DDL}(r, c, \hat{r}, \hat{c})}{\partial a_{ij}}a_{ij},
\end{equation*}
where $a_{ij}=Softmax(\hat{r}_{ij}-\lambda \hat{c}_{ij})$, and only $a$ is perturbed for gradient estimation and no longer for $\hat{r}, \hat{c}$.
\section{Evaluation}
In this section, we will conduct large-scale offline and online experiments to compare our methods with other benchmarks to validate their performance. 

\subsection{Offline Experiment}
\subsubsection{Dataset} Two types of datasets are provided in this paper.
\begin{itemize}[leftmargin=*]
\item \textbf{CRITEO-UPLIFT v2.} This public dataset is provided by the AdTech company Criteo in the AdKDD'18 workshop\cite{diemert2018large}. The dataset contains 13.9 million samples collected from a random control trial (RCT) that prevents a random part of users from being targeted by advertising. Each sample has 12 features, 1 binary treatment indicator and 2 response labels(visit/conversion). In order to study resource allocation problem under limited budget using the dataset, we follow\cite{zhou2023direct} and take the visit/conversion label as the cost/value respectively. We randomly sample 70\% samples for training and the remaining samples for test.

\item \textbf{Marketing data.} Discounting is a common marketing campaign in Meituan, an online food delivery platform. We conduct a two-week RCT to collect data in this platform. The online shops on the platform offer daily discounts to users. Note that to avoid price discrimination, the discount of a shop is the same for all individuals, but it changes randomly each day and varies from shop to shop. The data in the first week is used for training and the other for test. The discount $T \in \{0, 5, 10, 15, 20\}$ is taken as the treatment, where $T=t$ means $t\%$ off for each order whose price meets a given threshold. The dataset contains 2.8 million samples, and each sample has 107 features, 1 treatment label and 2 response labels (daily cost/orders).
\end{itemize}
\subsubsection{Evaluation Metrics}
Multiple evaluation metrics are provided for offline evaluation in this experiment. In addition to adopting the evaluation metrics commonly used in two-stage models, such as Logloss and MSE, we also use the following metrics for policy evaluation with counterfactuals, which are more significant.
\begin{itemize}[leftmargin=*]
\item \textbf{{AUCC (Area under Cost Curve).}} A common metric used in existing works~\cite{ai2022lbcf, du2019improve,zhou2023direct}, which is designed for evaluating the performance to rank ROI of individuals in the binary treatment setting. We use the metric to compare the performance of different methods in CRITEO-UPLIFT v2. 

\item \textbf{{EOM (Expected Outcome Metric).}} EOM is also commonly used in~\cite{ai2022lbcf, zhou2023direct, zhao2017uplift}. Based on RCT data, an unbiased estimation of the expected outcome (per-capita revenue/per-capita cost) for arbitrary budget allocation policy can be obtained. The details of EOM are shown in Sec.~\ref{subsec:FD}. We use the metric to compare the performance of different methods in Marketing data. 
\end{itemize}
\subsubsection{Benchmarks.}For each dataset in this paper, multiple models and algorithms are implemented and taken as benchmarks.
\begin{itemize}[leftmargin=*]
\item \textbf{TSM-SL.} The two-stage method is mentioned in many exsting works\cite{albert2022commerce, wang2023multi, zhao2019unified, ai2022lbcf}. In the first stage, a well-trained S-Learner model is used to predict the response (revenue/cost) of individuals under different treatments. In the second stage, we find the optimal budget allocation solution for an MCKP formulation based on the predictions.

\item \textbf{TSM-CF.} Also a two-stage method, the difference with TSM-SL is that instead of S-learner, we use a Causal Forests \cite{athey2019generalized} to predict the incremental response in the first stage. It is implemented here base on EconML packages \cite{battocchi2019econml}, which can support binary treatment and multiple treatments.

\item \textbf{DPM.} This method\cite{zhou2023direct} designs the decision factor for the MCKP, and proposes a surrogate loss to directly learn the decision factor.

\item \textbf{CN.} This method\cite{wang2023multi} imposes a monotonic constraint between outcome predictions and treatments, which is particularly useful for ITE estimation under multiple treatments. The method is trained with MSE loss and evaluated only on marketing data, which is a multi-treatment experiment. 

\item \textbf{CN+DFCL-PL.} The constraint network is trained with Decision-Focused Causal Learning (DFCL) loss, which comprises MSE loss ($\mathcal{L}_{PL}$) and policy learning loss ($\mathcal{L}_{PLL}$). 

\item \textbf{DFCL-PL.} The DFCL method based on policy learning loss proposed in this paper.

\item \textbf{DFCL-MER.} The DFCL method proposed in this paper utilizes the surrogate loss derived by Maximun Entropy Regularizer

\item \textbf{DFCL-IFD.} The DFCL method proposed in this paper for gradient estimation using the improved finite difference strategy.
\end{itemize}
\subsubsection{Implementation Details}
\begin{itemize}[leftmargin=*]
\item \textbf{CRITEO-UPLIFT v2.} For the baseline methods (TSM-SL, TSM-CF and DPM), we cite the results directly from\cite{zhou2023direct}. The DFCL model uses the same DNN architecture with a shared layer that is a single-layer MLP of dimension 128 and four head networks that are two-layer MLPs of dimension [64, 1]. Except for the final output layer, the remaining layers use ReLU activations. For DFCL-MER, we set the temperature $\tau=3$. Our models are trained for 40 epochs with the Adam optimizer \cite{kingma2014adam}. In order to accelerate the training, the first twenty epochs are warmstarting \cite{mandi2020smart} using the cross-entropy loss, and then the models are trained using the DFCL loss.

\item \textbf{Marketing data.} In the multi-treatment experiment, the models need to predict the revenue and cost under five treatments. TSM-SL, CN, CN+DFCL-PL, DFCL-PL, DFCL-MER and DFCL-IFD use the same DNN architecture: a 4 layers MLP (64-32-32-10). The first five outputs of the models are the predicted revenue, and the remaining outputs are the predicted cost. For DFCL-MER, we set the temperature $\tau=0.01$. For DPM, a S-Learner model is trained using the customized loss proposed in \cite{zhou2023direct} to directly predict marginal utility under different treatments. The DPM model has 4 layers of MLP (64-32-32-1), with the last layer using a sigmoid activation and each of the remaining layers with ReLU activations. All neural network-based models are trained for 500 epochs using the Adam optimizer. For TSM-CF, we set $n\_estimators=256$, $min\_sample\_leaf=300$ and $depth=24$.
\end{itemize}
All experiments are run on AMD EPYC 7502P Rome 32x@ 2.50GHz processor with 64GB memory.

\begin{table}[htbp]
\caption{Comparison of common metrics, noting that DPM and TSM-CF predict the decision factor and the incremental intervention effect, respectively, and thus do not apply to these two metrics.}
\label{tab:common metric}
\begin{tabular}{ccc}
\hline
\multirow{2}{*}{Model} & CRITEO-UPLIFT v2        & Marketing data          \\ \cline{2-3} 
                       & \multicolumn{1}{c}{Logloss} & \multicolumn{1}{c}{MSE} \\ \hline
TSM-SL                 &$0.2165\pm0.0001$                         & $0.2625\pm 0.0009$                        \\
CN                 &/                         &  $0.2639\pm0.0012$                        \\
CN+DFCL-PL                 &/                        &    $0.2703\pm 0.0015$           \\
DFCL-PL                &$0.2186\pm0.0008$                         & $0.2678\pm0.0010$                         \\
DFCL-MER               &$0.2178\pm0.0012$                         &$0.2650 \pm 0.0005$                         \\
DFCL-IFD               &$0.2170\pm0.0003$                         & $0.2642\pm 0.0009$                         \\ \hline
\end{tabular}
\end{table}

\begin{table}[htbp]
  \caption{\textbf{AUCC(CRITEO-UPLIFT v2)}}
  \label{tab:aucc}
    \begin{tabular}{ccc}
        \toprule[1pt]
        Model    & AUCC   & Improvement \\ \hline
        TSM-SL   & $0.7561\pm0.0113$ & /           \\
        TSM-CF   & $0.7558\pm0.0012$ & -0.03\%          \\
        DPM      & $0.7739\pm0.0002$ & +2.35\%           \\
        DFCL-PL  & $0.7713\pm0.0025$       & +2.01\%           \\
        DFCL-MER & $0.7727\pm0.0015$       & +2.20\%           \\
        DFCL-IFD & $0.7859\pm0.0021$       & +3.94\%           \\ \bottomrule[1pt]
        \end{tabular}
\end{table}

\begin{figure}[htbp]    
  \centering            
  \subfloat[AUCC(CRITEO-UPLIFT v2)]   
  {
      \label{fig:aucc}\includegraphics[width=0.48\columnwidth]{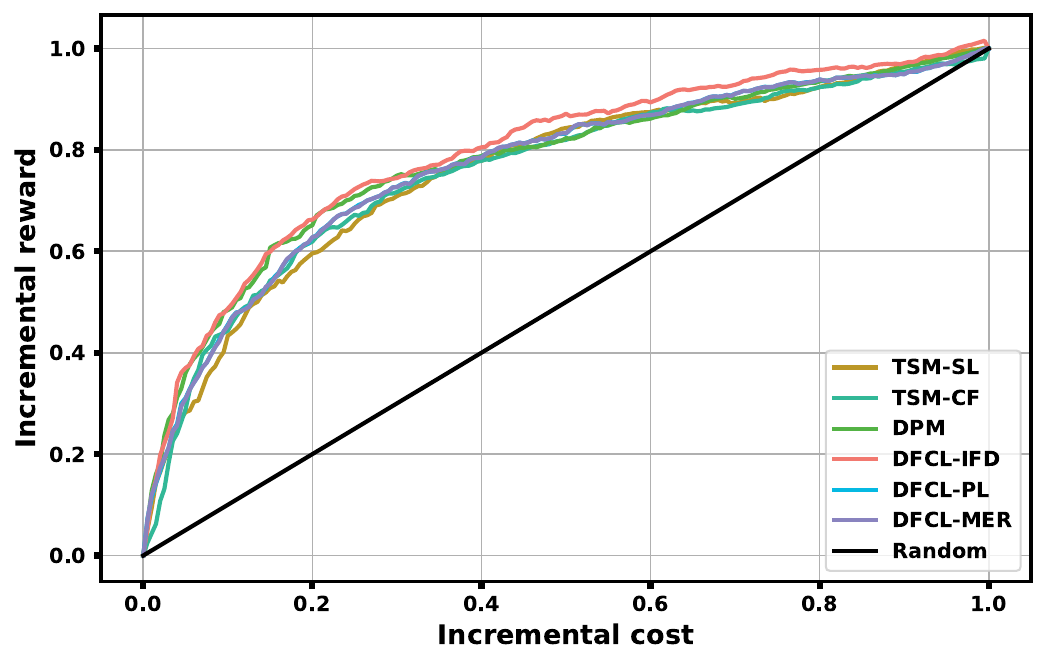}
  }
  \subfloat[EOM (Marketing data)]
  {
      \label{fig:eom}\includegraphics[width=0.49\columnwidth]{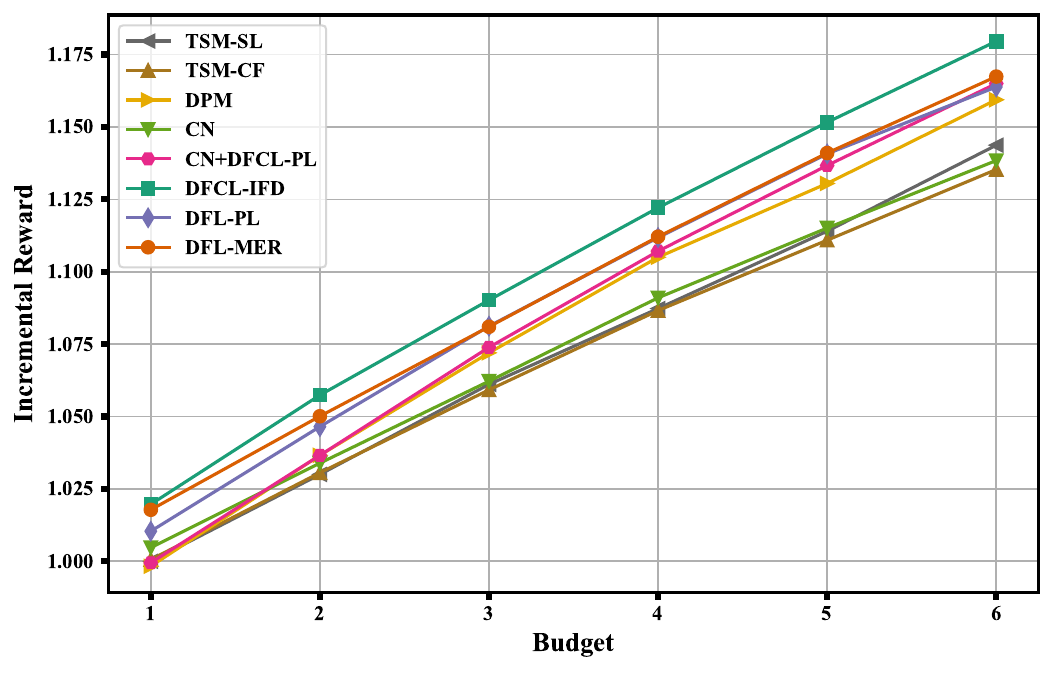}
  }
  \caption{Offline experiment results}    
  \label{fig:Offline_experiment_results}            
\end{figure}

\begin{table*}[htbp]
  \caption{\textbf{EOM (Marketing data)}}
  \label{tab:EOM}
\scalebox{0.9}{\begin{tabular}{cccccccc}
\toprule[1pt]
\multirow{2}{*}{Model} & \multicolumn{6}{c}{Budget}                          & \multirow{2}{*}{Improvement} \\ \cline{2-7}
                       & 1      & 2      & 3      & 4      & 5      & 6      &                              \\ \hline
TSM-SL                 & $1.0000\pm0.0023$ & $1.0300\pm0.0022$ & $1.0611\pm0.0023$ & $1.0873\pm0.0022$ & $1.1140\pm0.0020$ & $1.1437\pm0.0021$ &/              \\
TSM-CF                 & $1.0006\pm0.0007$ & $1.0306\pm0.0006$ & $1.0592\pm0.0004$ & $1.0866\pm0.0006$ & $1.1109\pm0.0003$ & $1.1353\pm0.0008$ & -0.19\%                       \\
DPM                    & $0.9983\pm0.0011$ & $1.0366\pm0.0010$ & $1.0720\pm0.0006$ & $1.1050\pm0.0003$ & $1.1305\pm0.0010$ & $1.1594\pm0.0007$ & 1.00\%                       \\
CN                    & $1.0047\pm0.0013$ & $1.0339\pm0.0010$ & $1.0622\pm0.0007$ & $1.0910\pm0.0009$ & $1.1151\pm0.0011$ & $1.1384\pm0.0015$ & 0.16\%                       \\
CN+DFCL-PL                    & $0.9995\pm0.0003$ & $1.0366\pm0.0008$ & $1.0739\pm0.0009$ & $1.1071\pm0.0005$ & $1.1367\pm0.0006$ & $1.1650\pm0.0009$ & 1.26\%                       \\
DFCL-PL                & $1.0104\pm0.0005$ & $1.0465\pm0.0006$ & $1.0812\pm0.0004$ & $1.1118\pm0.0007$ & $1.1407\pm0.0011$ & $1.1638\pm0.0018$ & 1.98\%                       \\
DFCL-MER               & $1.0178\pm0.0008$ & $1.0501\pm0.0005$ & $1.0810\pm0.0002$ & $1.1121\pm0.0010$ & $1.1410\pm0.0013$ & $1.1674\pm0.0009$ & 2.06\%                       \\
DFCL-IFD               & $1.0197\pm0.0012$ & $1.0574\pm0.0022$ & $1.0902\pm0.0024$ & $1.1221\pm0.0026$ & $1.1516\pm0.0028$ & $1.1796\pm0.0030$ & 2.85\%                       \\ \bottomrule[1pt]
\end{tabular}}
\end{table*}

\subsection{Experimental Results}
\subsubsection{Overall performance}
In Table \ref{tab:common metric}, we present the prediction loss of different models on the two datasets. Clearly, the two-stage method performs best on common metrics, which minimizes MSE or Logloss on the training set. However, what we really focus on is the decision quality of predictions. Fig.~\ref{fig:aucc} and Table~\ref{tab:aucc} present the comparison between our proposed methods and other benchmarks in CRITEO-UPLIFT v2 on AUCC~\cite{du2019improve}, which represents the decision quality under binary treatments. We can see that DFCL-IFD achieves the best performance, DFCL-PL, and DFCL-MER perform similarly to DPM, and the two-stage methods perform the worst. 

In marketing data, we use EOM method to calculate per-capita orders and per-capita budgets based on predictions. The results are shown in Table~\ref{tab:EOM} and Fig. \ref{fig:eom}. Our models significantly outperform the baseline models in terms of per-capita orders at all per-capita budgets. DPM is on par with the two stage methods in the low per-capita budgets and outperforms them in the high per-capita budgets. CN has a marginal improvement of 0.16\% compared to the two-stage methods. Further evaluation is carried out on the model trained with DFCL loss, which comprises MSE loss ($\mathcal{L}_{PL}$) and policy learning loss ($\mathcal{L}_{PLL}$). The integration of policy learning loss yielded a notable enhancement in performance, with the constrained network showing a significant increase of 1.26\%. These findings suggest that our proposed DFCL approach is versatile and can be integrated into existing methodologies. Interestingly, the constraint network combined with policy learning loss (CN + DFCL-PL) did not outperform DFCL-PL alone. We hypothesize that this may be due to the predefined constraints within the network, which potentially restrict the expansiveness of the decision space.
\subsubsection{Prediction Loss vs Decision Loss tradeoff}
As mentioned above, we integrate the prediction loss as a regularizer into the training objective. In this experiment, we will consider how the weight of the prediction loss affects the performance of DFCL. We set $\alpha \in \{0.1, 0.5, 1, 2, 3, 4, 5, 10\}$ and measure the per-capita orders under a fixed per-capita budget. As shown in Fig. \ref{fig:alpha}, increasing $\alpha$ in a certain range does not lead to a decrease in model performance. However, if $\alpha$ is too large, the prediction loss dominates the training objective and the model will be reduced to the two-stage method. The experiment suggests it is possible to choose a value of $\alpha$ so that we can achieve better performance and more accurate predictions.

\subsubsection{Impact of Lagrange multiplier}
\begin{figure}[h]   
  \centering            
    \subfloat[Impact of the prediction loss weight $\alpha$]
  {
      \label{fig:alpha}\includegraphics[width=0.48\columnwidth]{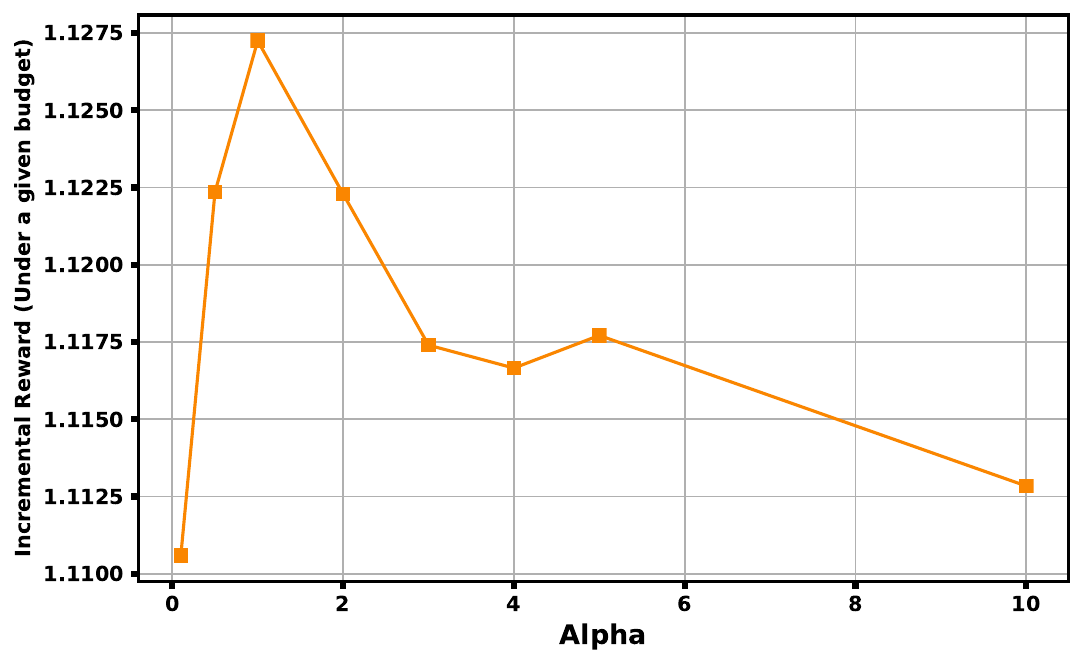}
  } 
    \subfloat[Impact of Lagrange multiplier]
  {
      \label{fig:lambda}\includegraphics[width=0.48\columnwidth]{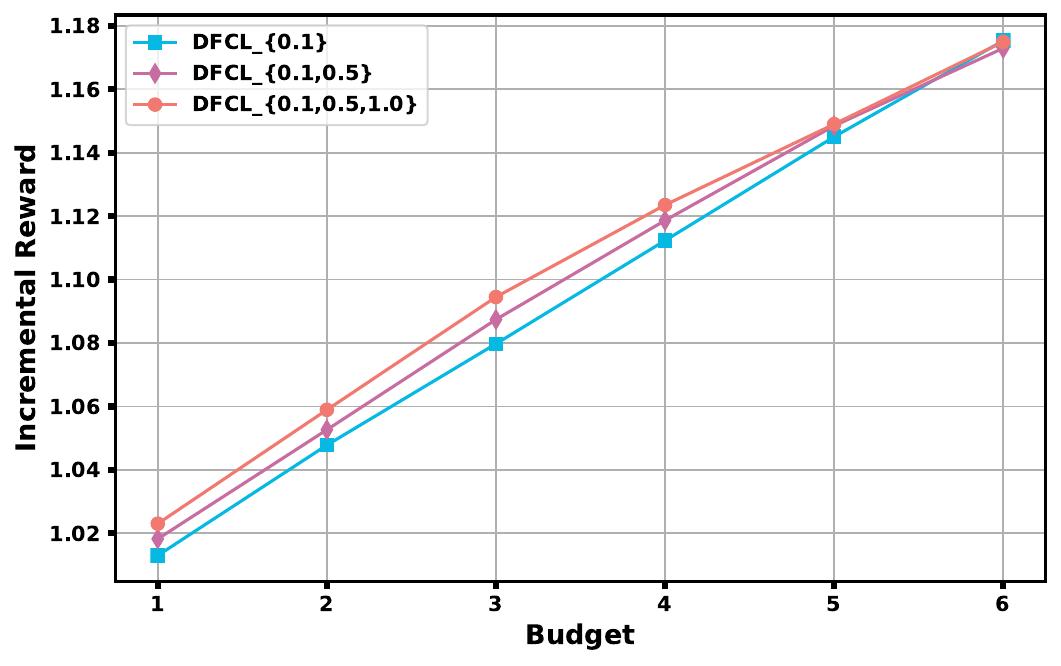}
  }
  \caption{Offline experiment results}    
  \label{fig:Offline_experiment_results}            
\end{figure}

Next, we would like to discuss the impact of the Lagrange multiplier $\lambda$ on the performance of DFCL model. Since a given Lagrange multiplier $\lambda$ corresponds to the MCKP for a certain budget constraint, DFCL model can learn allocation policies for different budgets simultaneously by changing or adding $\lambda$ to the DFCL loss. We set up different combinations of Lagrange multipliers ( $\lambda \in \{\{0.1\}, \{0.1, 0.5\}, \{0.1, 0.5, 1.0\}\}$) and use to train DFCL models. Fig. \ref{fig:lambda} shows the results using DFCL-IFD models trained by combinations of Lagrange multipliers. We can observe that $\lambda$ is a hyperparameter that can have a significant impact on model performance. A small $\lambda$ enables the model to learn the allocation policy efficiently under high budget and vice versa. Moreover, models trained with multiple Lagrange multipliers can balance performance with different budgets.

\subsection{Online A/B testing}
\subsubsection{Setups}
\begin{figure}[htbp]    
  \centering            
    \subfloat[Online deployment of DFCL]
  {
      \label{fig:deployment}\includegraphics[width=0.48\columnwidth]{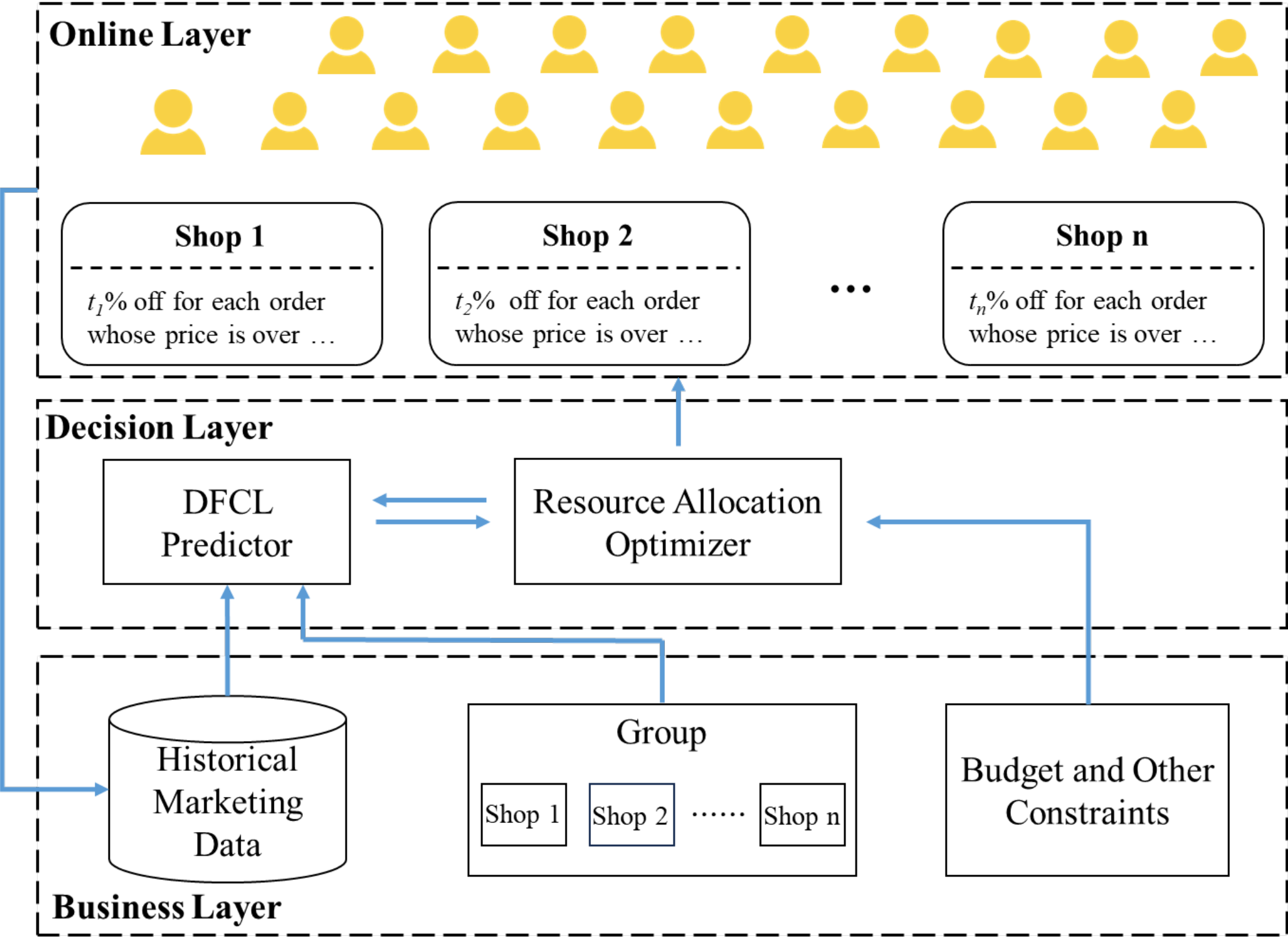}
  } 
    \subfloat[Orders]
  {
      \label{fig:online test}\includegraphics[width=0.5\columnwidth]{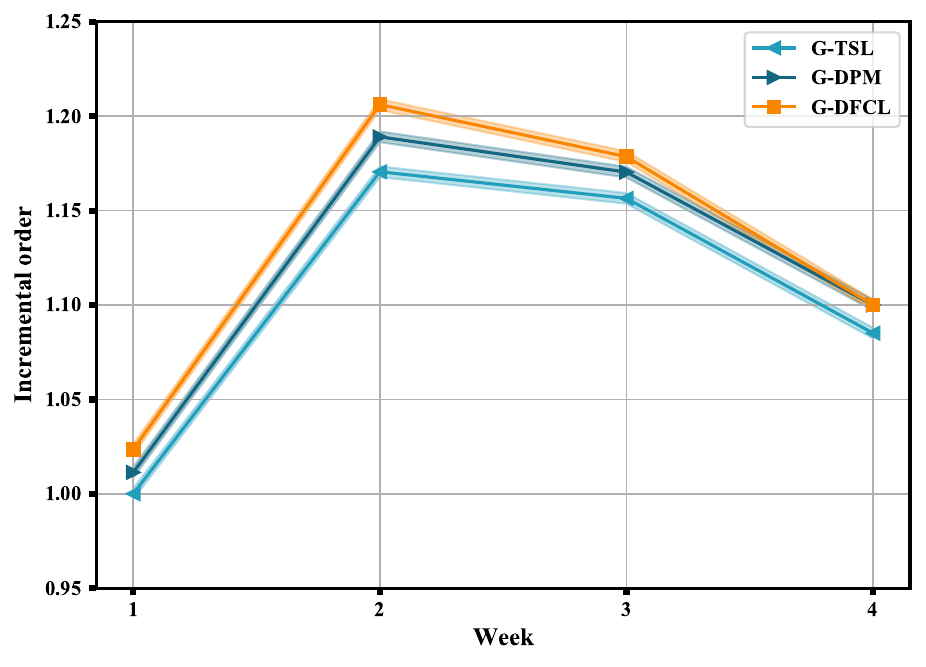}
  }
  \caption{Online A/B testing}    
  \label{fig:Offline_experiment_results}            
\end{figure}

We deploy DFCL, DPM and TSM-SL to support a discount campaign in Meituan (a food delivery platform), and conduct an online A/B testing for four weeks. The experiment contains 310K online shops and they are randomly divided every day into three groups called G-DFCL, G-DPM and G-TSL respectively. Each shop will be assigned a discount $t \in \{0, 5, 10, 15, 20\}$ as the treatmemt, which means $t\%$ off for each order whose price meets a given threshold. The marketing goal is to maximize the orders by assigning an appropriate discount to each store every day for a limited budget. The online deployment of DFCL is shown in Fig. \ref{fig:deployment}: (1) Before the campaign starts each day, we use the DFCL model to make predictions and allocate the appropriate discounts to each store based on budget and other constraints in an offline environment. (2) The users visit the online shop and get discounts which will stimulate them to make purchases. (3) During model training, we use historical random data and resource allocation optimizer to update the model parameters.

\subsubsection{Results} Fig.~\ref{fig:online test} illustrates the improvement in weekly orders for G-DFCL and G-DPM relative to G-TSM. In order to preserve data privacy, all data points in Fig.~\ref{fig:online test} have been normalized that are divided by the orders of TSM-SL in the first week. We can see that DFCL achieves a significant average improvement of 2.17\% relative to TSM-SL and also outperforms DPM with a relative improvement of 0.85\%. The detailed results can be found in Appendix~\ref{sec:detailed results}.
\section{Conclusion}
In this paper, we propose a decision focused causal learning framework (DFCL) for direct counterfactual marketing optimization, which overcomes the technological challenges of DFL deployment in marketing. By designing surrogate losses and constructing black-box optimisation, we efficiently align the objectives of ML and OR. Both offline experiments and online A/B testing demonstrate the effectiveness of DFCL over the state-of-the-art methods.

\begin{acks}
This work was supported in part by National Key R\&D Program of China (2023YFB4502400), 
the NSF of China (62172206), and the Xiaomi Foundation.
\end{acks}

\bibliographystyle{ACM-Reference-Format}
\bibliography{reference}


\appendix

\section{The Proof of Theorem~\ref{theorem:L_PL}}
\label{sec:proof_L_PL}
\begin{proof}
First of all, we introduce some notations. Following the potential outcome framework~\cite{sekhon2008neyman}, let $X \in \mathbb{R}^d$ denote the feature vector and $T \in \{1,2,\ldots,M\}$ be the treatment. Let $Y^r (T)$ and $Y^c(T)$ be the potential outcome of the revenue and the cost respectively when the individual receives treatment $T$. let $\widehat{Y}^r (T)$ and $\widehat{Y}^c(T)$ be the predicted outcome of the revenue and the cost respectively when the individual receives treatment $T$. 
For $\mathcal{L}_{MSE}$, we have
\begin{align*}
\mathcal{L}_{MSE}(r,c,\hat{r},\hat{c}) &= \frac{1}{NM} \sum_{i} \sum_{j} (r_{ij}-\hat{r}_{ij})^{2}+(c_{ij}-\hat{c}_{ij})^{2}\\
&= \mathbb{E}_{X,T} [(Y^r(T)-\widehat{Y}^r(T))^{2}+(Y^c(T)-\widehat{Y}^c(T))^{2}].
\end{align*}
For $\mathcal{L}_{PL}$, we have
\begin{align*}
&\mathcal{L}_{PL}(r,c,\hat{r},\hat{c})  \\
=& \frac{1}{M}\sum_{i}\frac{1}{N_{t_i}} [(r_{it_{i}}-\hat{r}_{it_{i}})^{2}+(c_{it_{i}}-\hat{c}_{it_{i}})^{2}] \\
=& \frac{1}{M}\sum_{j}\sum_{i:t_i=j}\frac{1}{N_{t_i}} [(r_{it_{i}}-\hat{r}_{it_{i}})^{2}+(c_{it_{i}}-\hat{c}_{it_{i}})^{2}] \\
=& \frac{1}{M}\sum_{j}\frac{1}{N_{t_i}} \sum_{i:t_i=j} [(r_{it_{i}}-\hat{r}_{it_{i}})^{2}+(c_{it_{i}}-\hat{c}_{it_{i}})^{2}] \\
=& \frac{1}{M}\sum_{j} \mathbb{E}_X[(Y^r(j)-\widehat{Y}^r(j))^{2}+(Y^c(j)-\widehat{Y}^c(j))^{2} | T_i = j] \\
=& \frac{1}{M}\sum_{j} \mathbb{E}_X[(Y^r(j)-\widehat{Y}^r(j))^{2}+(Y^c(j)-\widehat{Y}^c(j))^{2}] \ \ \ \ \ \ (T \perp X) \\
=& \mathbb{E}_{X,T} [(Y^r(T)-\widehat{Y}^r(T))^{2}+(Y^c(T)-\widehat{Y}^c(T))^{2}],
\end{align*}
where $T\perp X$ holds because the data set is from random control trials (RCT). (RCT).
Therefore, we finish the proof.
\end{proof}

\begin{algorithm}
	\renewcommand{\algorithmicrequire}{\textbf{Input:}}
	\renewcommand{\algorithmicensure}{\textbf{Output:}}
	\caption{Lagrangian duality gradient estimator}
	\label{alg:gradient estimator}
	\begin{algorithmic}[1]
		\REQUIRE training data D $\equiv \{(x_{i}, t_{i}, r_{it_{i}}, c_{it_{i}})\}^{N}_{i=1}$, Lagrange multiplier $\lambda$, the predicted revenue/cost $\hat{r}/\hat{c}$
            \ENSURE $\frac{\partial \mathcal{L}_{DDL}(\lambda, r, c, \hat{r}, \hat{c})}{\partial \hat{r}}$, $\frac{\partial \mathcal{L}_{DDL}(\lambda, r, c, \hat{r}, \hat{c})}{\partial \hat{c}}$
		\STATE Initialize $\frac{\partial\mathcal{L}_{DDL}(\lambda, r, c, \hat{r}, \hat{c})}{\partial \hat{r}}=0$, $\frac{\partial \mathcal{L}_{DDL}(\lambda, r, c, \hat{r}, \hat{c})}{\partial \hat{c}}=0$, $z_{ij}=0 \ \forall i,j$
            \STATE $a=\hat{r}-\lambda\hat{c}$
            \STATE $\forall i, j, \ z_{ij} = \mathbb{I}_{j = \arg \max_j (a_{ij})}$
            \STATE $\bar{r}(\hat{r}, \hat{c}, \lambda)=\frac{1}{N} \sum_{i} \frac{1}{p_{t_{i}}} r_{t_{i}} \mathbb{I}_{t_{i}=\arg\max_{j} z_{ij}}$
            \STATE $\bar{c}(\hat{r}, \hat{c}, \lambda)=\frac{1}{N} \sum_{i} \frac{1}{p_{t_{i}}} c_{t_{i}} \mathbb{I}_{t_{i}=\arg\max_{j} z_{ij}}$
            \STATE $\mathcal{L}_{DDL}(\lambda, r, c, \hat{r}, \hat{c})=\bar{r}(\lambda, r, c, \hat{r}, \hat{c}) - \lambda \bar{c}(\lambda, r, c, \hat{r}, \hat{c})$
            \STATE matching\_indices= $\{i|t_{i}=\arg\max_{j} \ z_{ij}, \ \forall i \}$
            \STATE mismatching\_indices= $\{i|t_{i}\neq \arg\max_{j} \ z_{ij}, \ \forall i \}$
            \FORALL{$i \in$ matching\_indices}
                \STATE $h^{r}_{it_{i}}=max_{j \neq t_{i}}a_{ij}-a_{it_{i}}$, $h^{c}_{it_{i}}=\frac{(a_{it_{i}}-\max_{j \neq t_{i}}a_{ij})}{\lambda}$
                \STATE $\frac{\partial\mathcal{L}_{DDL}(\lambda, r, c, \hat{r}, \hat{c})}{\partial \hat{r}_{it_{i}}}=\frac{-\frac{1}{Np_{t_{i}}}(r_{it_{i}}-\lambda c_{it_{i}})}{h^{r}_{it_{i}}}$
                \STATE $\frac{\partial \mathcal{L}_{DDL}(\lambda, r, c, \hat{r}, \hat{c})}{\partial \hat{c}_{it_{i}}}=\frac{-\frac{1}{Np_{t_{i}}}(r_{it_{i}}-\lambda c_{it_{i}})}{h^{c}_{it_{i}}}$
                \FORALL{$j \in \{1, 2, ..., M\} \ and \ j \neq t_{i}$}
                \STATE $h^{r}_{ij}=a_{it_{i}}-a_{ij}$, $h^{c}_{ij}=\frac{(a_{ij}-a_{it_{i}})}{\lambda}$ 
                \STATE $\frac{\partial \mathcal{L}_{DDL}(\lambda, r, c, \hat{r}, \hat{c})}{\partial \hat{r}_{ij}}=\frac{-\frac{1}{Np_{t_{i}}}(r_{it_{i}}-\lambda c_{it_{i}})}{h^{r}_{ij}}$
                \STATE $\frac{\partial \mathcal{L}_{DDL}(\lambda, r, c, \hat{r}, \hat{c})}{\partial \hat{c}_{ij}}=\frac{-\frac{1}{Np_{t_{i}}}(r_{it_{i}}-\lambda c_{it_{i}})}{h^{c}_{ij}}$
                \ENDFOR
            \ENDFOR
            \FORALL{$i \in$ mismatching\_indices}
                \STATE $j=\arg\max_{j} \ a_{ij}$
                \STATE $h^{r}_{it_{i}}=a_{ij}-a_{it_{i}}$, $h^{r}_{ij}=-h^{r}_{it_{i}}$
                \STATE $h^{c}_{it_{i}}=\frac{(max_{j}a_{ij}-a_{it_{i}})}{\lambda}$, $h^{c}_{ij}=-h^{c}_{it_{i}}$
                \STATE $\frac{\partial \mathcal{L}_{DDL}(\lambda, r, c, \hat{r}, \hat{c})}{\partial \hat{r}_{it_{i}}}=\frac{\frac{1}{Np_{t_{i}}}(r_{it_{i}}-\lambda c_{it_{i}})}{h^{r}_{it_{i}}}$
                \STATE $\frac{\partial \mathcal{L}_{DDL}(\lambda, r, c, \hat{r}, \hat{c})}{\partial \hat{c}_{it_{i}}}=\frac{\frac{1}{Np_{t_{i}}}(r_{it_{i}}-\lambda c_{it_{i}})}{h^{c}_{it_{i}}}$
                \STATE $\frac{\partial \mathcal{L}_{DDL}(\lambda, r, c, \hat{r}, \hat{c})}{\partial \hat{r}_{ij}}=\frac{\frac{1}{Np_{t_{i}}}(r_{it_{i}}-\lambda c_{it_{i}})}{h^{r}_{ij}}$
                \STATE $\frac{\partial \mathcal{L}_{DDL}(\lambda, r, c, \hat{r}, \hat{c})}{\partial \hat{c}_{ij}}=\frac{\frac{1}{Np_{t_{i}}}(r_{it_{i}}-\lambda c_{it_{i}})}{h^{c}_{ij}}$
            \ENDFOR
		
            \RETURN $\frac{\partial \mathcal{L}_{DDL}(\lambda, r, c, \hat{r}, \hat{c})}{\partial \hat{r}}$, $\frac{\partial \mathcal{L}_{DDL}(\lambda, r, c, \hat{r}, \hat{c})}{\partial \hat{c}}$
	\end{algorithmic}  
\end{algorithm}

\section{The Proof of Theorem~\ref{theorem:L_PLL}}
\label{sec:proof_L_PLL}
\begin{proof}
Follow the notations in Appendix~\ref{sec:proof_L_PL} and let
$$\text{softmax}(\hat{r}_{ij}-\lambda\hat{c}_{ij}) = \frac{\exp(\hat{r}_{ij} - \lambda \hat{c}_{ij})}{ \sum_k \exp(\hat{r}_{ik} - \lambda \hat{c}_{ik})}$$
be the softmax function.
Hence, $\mathcal{L}'_{DDL}$ can be rewritten as
\begin{align*}
&\mathcal{L}'_{DDL}(r,c,\hat{r},\hat{c}) \\
=& -\sum_{\lambda} \sum_i\sum_j (r_{ij} - \lambda c_{ij}) \text{softmax}(\hat{r}_{ij}-\lambda\hat{c}_{ij}) \\
=& -NM \sum_{\lambda} \frac{1}{NM}\sum_i\sum_j (r_{ij} - \lambda c_{ij}) \text{softmax}(\hat{r}_{ij}-\lambda\hat{c}_{ij}) \\
=& - NM\sum_{\lambda} \mathbb{E}_{X,T}[(Y^r(T) - \lambda Y^c(T)) \text{softmax}(\widehat{Y}^r(T) - \lambda \widehat{Y}^c(T))]
\end{align*}
In addition, we have
\begin{align*}
&\mathcal{L}_{PLL}(r,c,\hat{r},\hat{c}) \\
=&-\sum_{\lambda} \sum_{i}\frac{N}{N_{t_i}}(r_{it_{i}}-\lambda c_{it_{i}}) \text{softmax}(\hat{r}_{it_i}-\lambda\hat{c}_{it_i}) \\
=&-\sum_{\lambda} \sum_{i}\frac{N}{N_{t_i}}(r_{it_{i}}-\lambda c_{it_{i}}) \text{softmax}(\hat{r}_{it_i}-\lambda\hat{c}_{it_i}) \\
=&-N\sum_{\lambda} \sum_{j}\sum_{i:t_i=j}\frac{1}{N_{t_i}}(r_{it_{i}}-\lambda c_{it_{i}}) \text{softmax}(\hat{r}_{it_i}-\lambda\hat{c}_{it_i}) \\
=&-N\sum_{\lambda} \sum_{j} \mathbb{E}_{X}[(Y^r(j) - \lambda Y^c(j)) \text{softmax}(\widehat{Y}^r(j) - \lambda \widehat{Y}^c(j)) | T_i = j] \\
=& -N\sum_{\lambda} \sum_{j} \mathbb{E}_{X}[(Y^r(j) - \lambda Y^c(j)) \text{softmax}(\widehat{Y}^r(j) - \lambda \widehat{Y}^c(j))]\\
=& -NM\sum_{\lambda} \frac{1}{M}\sum_{j} \mathbb{E}_{X}[(Y^r(j) - \lambda Y^c(j)) \text{softmax}(\widehat{Y}^r(j) - \lambda \widehat{Y}^c(j))]\\
=& - NM\sum_{\lambda} \mathbb{E}_{X,T}[(Y^r(T) - \lambda Y^c(T)) \text{softmax}(\widehat{Y}^r(T) - \lambda \widehat{Y}^c(T))].
\end{align*}
Therefore, $\mathcal{L}'_{DDL}(r,c,\hat{r},\hat{c}) = \mathcal{L}_{PLL}(r,c,\hat{r},\hat{c})$ holds.

For $\forall i, j = \arg\max_k r_{ik} - \lambda c_{ik}$, let $\hat{r}_{ij}-\lambda \hat{c}_{ij} \to +\infty$; for $\forall i, j \neq \arg\max_k r_{ik} - \lambda c_{ik}$, let $\hat{r}_{ij}-\lambda \hat{c}_{ij} \to -\infty$. Hence, we have
$$\text{softmax}(\hat{r}_{ij}-\lambda\hat{c}_{ij}) \to \mathbb{I}_{j=\arg\max_k \hat{r}_{ik} - \lambda \hat{c}_{ik}}.$$
Therefore, we further get
$$\min_{\hat{r},\hat{c}} \mathcal{L}_{PLL}(\lambda,\hat{r},\hat{c}) = \min_{\hat{r},\hat{c}} \mathcal{L}'_{DDL}(\lambda,\hat{r},\hat{c}) = \min_{\hat{r},\hat{c}} \mathcal{L}_{DDL}(\lambda,\hat{r},\hat{c}).$$
\end{proof}

\begin{algorithm}
	\renewcommand{\algorithmicrequire}{\textbf{Input:}}
	\renewcommand{\algorithmicensure}{\textbf{Output:}}
	\caption{An implementation of per capita revenue estimation for primal MCKP with budget $B$}
	\label{alg:evaluation}
	\begin{algorithmic}[1]
		\REQUIRE training data D $\equiv \{(x_{i}, t_{i}, r_{it_{i}}, c_{it_{i}})\}^{N}_{i=1}$, the budget $B$, the predicted revenue/cost $\hat{r}/\hat{c}$, a small constant $\epsilon$
            \ENSURE the expected per capita revenue $\bar{r}(B, r, c, \hat{r}, \hat{c})$
		\STATE Initialize $\lambda_{\min}=0$, $\lambda_{\max}=\max_{i,j}(\frac{\hat{r_{ij}}}{\hat{c_{ij}}})$, $z_{ij}=0 \ \forall i,j$
		\WHILE{$\frac{B}{N}-\bar{c}(\lambda, r, c, \hat{r}, \hat{c})<\epsilon$}
                \STATE $\lambda = \frac{\lambda_{\max} + \lambda_{\min}}{2}$
                \STATE $\forall i, j, \ z_{ij} = \mathbb{I}_{j = \arg \max_j (\hat{r_{ij}}-\lambda \hat{c_{ij}})}$
		      \STATE $\bar{r}(\lambda, r, c, \hat{r}, \hat{c})=\frac{1}{N} \sum_{i} \frac{1}{p_{t_{i}}} r_{t_{i}} \mathbb{I}_{t_{i}=\arg\max_{j} \ z_{ij}}$
                \STATE $\bar{c}(\lambda, r, c, \hat{r}, \hat{c})=\frac{1}{N} \sum_{i} \frac{1}{p_{t_{i}}} c_{t_{i}} \mathbb{I}_{t_{i}=\arg\max_{j} \ z_{ij}}$
                \IF{$\frac{B}{N}-\bar{c}(\lambda, r, c, \hat{r}, \hat{c})>0$}
                    \STATE $\lambda_{\max}=\lambda$
                \ELSE
                    \STATE $\lambda_{\min}=\lambda$
                \ENDIF
		\ENDWHILE
            \STATE $\bar{r}(B, r, c, \hat{r}, \hat{c})$ = $\bar{r}(\lambda, r, c, \hat{r}, \hat{c})$
            \RETURN the expected per capita revenue $\bar{r}(B, r, c, \hat{r}, \hat{c})$
	\end{algorithmic}  
\end{algorithm}

\begin{table*}[htbp]
\caption{\textbf{Online A/B testing results with the confidence interval}}
\label{tab:A/B testing}
\scalebox{1}{
\begin{tabular}{cccccc}
        \toprule[1pt]
        \multirow{2}{*}{Group} & \multicolumn{4}{c}{Week}          & \multirow{2}{*}{Improvement} \\ \cline{2-5}
                               & 1st      & 2nd      & 3rd      & 4th      &                              \\ \hline
        G-TSL                  & $1.0000\pm0.00285$  & $1.1706\pm 0.00298$ & $1.1565\pm0.00293$ & $1.0851\pm 0.00289$ & /                            \\
        G-DPM                  & $1.0113\pm 0.00284$ & $1.1891 \pm 0.00298$ & $1.1704\pm 0.00293$ & $1.1000\pm0.00288$ & 1.32\%                       \\
        G-DFCL                 & $1.0235\pm0.00285$ & $1.2062\pm0.00297$ & $1.1786\pm0.00293$ & $1.1000\pm0.00288$ & 2.17\%                       \\ \bottomrule[1pt]
\end{tabular}
}
\end{table*}

\section{Policy Evaluation Based on EOM}
\label{sec:EOM}
Given a batch of $N$ random samples and model predictions $\hat{r}$ and $\hat{c}$, we can use binary search to empirically estimate the per capita revenue under a per capita budget $\frac{B}{N}$, Algorithm \ref{alg:evaluation} summarizes this approach. 

\section{Lagrangian Duality Gradient Estimator}
\label{sec:estimator}
The decision making is independent for each individual thanks to the decomposition of the Lagrangian duality theory. Thus, for each sample, the smallest perturbation that causes a change in the dual decison loss is first calculated, and the loss after the perturbation is obtained by correcting only the original result. Algorithm \ref{alg:gradient estimator} provides details of the modified gradient estimator, which greatly reduces the computational overhead. Note that for comprehensibility, Algorithm \ref{alg:gradient estimator} is described with for loops, while in practice we work with matrix operations.

\section{Supplementary Experimental Results}
\label{sec:detailed results}
Tabel~\ref{tab:A/B testing} presents the detailed online A/B testing results. In order to preserve data privacy, all data points have been normalized by dividing by the orders of TSM-SL in the first week. The confidence interval ($\alpha=0.05$) is computed by a t-test.

\end{document}